\documentclass[runningheads]{llncs}

\usepackage{eccv}

\usepackage{eccvabbrv}

\usepackage{graphicx}
\usepackage{booktabs}

\usepackage[accsupp]{axessibility}  % Improves PDF readability for those with disabilities.

\usepackage{hyperref}

\usepackage{orcidlink}
\usepackage{bm}
\newcommand\norm[1]{\lVert#1\rVert}
\usepackage{amsmath}
\usepackage{amssymb}
\usepackage{graphicx}
\usepackage{bbding}
\usepackage{booktabs}
\usepackage{pifont}
\usepackage{multirow}
\usepackage{wrapfig}
\newcommand{\xmark}{\ding{55}}%

\begin{document}

\title{Rethinking Image-to-Video Adaptation: An Object-centric Perspective} 

\titlerunning{Object-centric Image-to-Video Adaptation}

\author{Rui Qian\inst{1}\orcidlink{0000-0002-0378-6438} \and
Shuangrui Ding\inst{1}\orcidlink{0000-0001-7033-774X} \and
Dahua Lin\inst{1,2}\thanks{Corresponding author.}\orcidlink{0000-0002-8865-7896}}

\authorrunning{R.~Qian et al.}

\institute{The Chinese University of Hong Kong, Hong Kong, China \and
Shanghai Artificial Intelligence Laboratory, Shanghai, China
\email{\{qr021,ds023,dhlin\}@ie.cuhk.edu.hk}}

\maketitle

\begin{abstract}
  Image-to-video adaptation seeks to efficiently adapt image models for use in the video domain. Instead of finetuning the entire image backbone, many image-to-video adaptation paradigms use lightweight adapters for temporal modeling on top of the spatial module. However, these attempts are subject to limitations in efficiency and interpretability. In this paper, we propose a novel and efficient image-to-video adaptation strategy from the object-centric perspective. Inspired by human perception, which identifies objects as key components for video understanding, we integrate a proxy task of object discovery into image-to-video transfer learning. Specifically, we adopt slot attention with learnable queries to distill each frame into a compact set of object tokens. These object-centric tokens are then processed through object-time interaction layers to model object state changes across time. Integrated with two novel object-level losses, we demonstrate the feasibility of performing efficient temporal reasoning solely on the compressed object-centric representations for video downstream tasks. Our method achieves state-of-the-art performance with fewer tunable parameters, only 5\% of fully finetuned models and 50\% of efficient tuning methods, on action recognition benchmarks. In addition, our model performs favorably in zero-shot video object segmentation without further retraining or object annotations, proving the effectiveness of object-centric video understanding.
\end{abstract}

\section{Introduction}
\label{sec:intro}
Recently, large-scale image pre-training has revolutionized the computer vision community, with the large foundation models dominating various image tasks~\cite{radford2021learning,yuan2021florence,jia2021scaling,wang2022image,oquab2023dinov2,zhou2021ibot}. The remarkable success of these models can be attributed to the enormous amounts of image data~\cite{oquab2023dinov2,he2022masked} or image-text pairs~\cite{radford2021learning,wang2022image,jia2021scaling}.  
\begin{figure}
    \centering
    \includegraphics[width=0.88\linewidth]{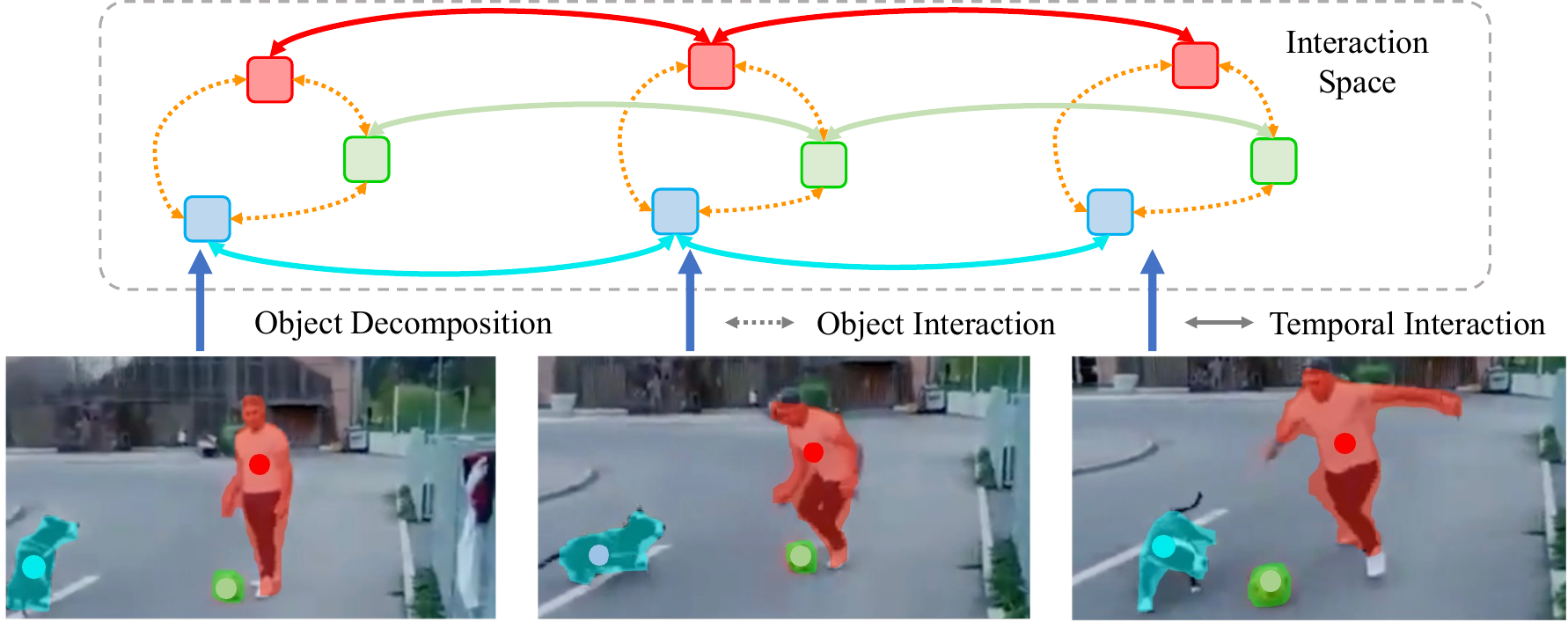}
    \caption{Illustration of our proposed object-centric video understanding pipeline for image-to-video adaptation. Specifically, we first parse each frame into several object components (represented in different colors) to form an interaction space. Then we respectively establish inter-object interactions within each frame (represented by dotted arrows), and temporal state changes of individual objects (depicted by solid arrows).}
     % concept of object-centric video analysis, where a video is processed as a series of interactions among a set of objects. 
    \label{teaser}
\end{figure}
However, when we extend this discussion to the video domain, it becomes non-trivial. Acquiring large video datasets and managing high computational costs make training video foundation models from scratch a significant challenge.
To circumvent these difficulties, the community has developed an innovative concept dubbed image-to-video adaptation. This approach aims to leverage the power of pre-trained image models and apply them to video data, enhancing both performance and efficiency. 
The image-to-video adaptation establishes a connection between effective methodologies in image models and the realm of video understanding, thereby opening new avenues for research and exploration.

% The standard image-to-video training paradigms are to first initialize from an image pre-trained model, then introduce architectural modifications to enable temporal reasoning, and finally finetune the model on video data. 
% % 
% Contrasting with the computationally expensive fully fine-tuning paradigm employed in previous work~\cite{bertasius2021space,carreira2017quo,lin2019tsm,arnab2021vivit,liu2022video,ni2022expanding}, 
Motivated by parameter-efficient transfer learning in natural language processing~\cite{houlsby2019parameter,hu2021lora,zaken2021bitfit,lester2021power},
a series of recent studies~\cite{pan2022st,yang2023aim,ju2022prompting,ni2022expanding,lin2022frozen,liu2023revisiting} have adopted an efficient tuning approach for image-to-video adaptation. These approaches maintain the image pre-trained model in a frozen state and only finetune a minimal set of additional parameters.
For example, \cite{pan2022st,yang2023aim,ni2022expanding,liu2023revisiting} insert learnable spatio-temporal modules into Transformer encoder blocks. 
% \cite{ju2022prompting,lin2022frozen} directly attach extra spatio-temporal aggregation layers to the frame-wise features. 
However, this formulation encounters two vital limitations. Firstly, tuning the in-between parameters in the shallow layers still requires a considerable amount of GPU memory during gradient backpropagation. 
Secondly, the introduction of basic temporal modules, like attention layers, makes temporal reasoning on highly redundant features and lacks the essential inductive bias to effectively benefit the image-to-video adaptation, thereby presenting challenges in interpreting the training's success.

To tackle these challenges, we propose to transfer object knowledge from frozen image pre-trained models to videos for more efficient image-to-video adaptation. In essence, our approach entails two main steps: decoupling each frame into distinct object components, and subsequently identifying object state changes across time to facilitate comprehensive video comprehension as illustrated in Fig.~\ref{teaser}.
% Specifically, inspired by recent works on object discovery~\cite{locatello2020object,xu2022groupvit,xie2022segmenting}, we use a compact set of learnable query tokens to decompose different objects and perform temporal reasoning on these object tokens as illustrated in Fig.~\ref{teaser}. 
Specifically, we first pass the video frames through the image pre-trained model to obtain frame-wise features. Then, inspired by recent works on object discovery~\cite{locatello2020object,xu2022groupvit,xie2022segmenting}, we employ slot attention~\cite{locatello2020object} with learnable queries~\cite{Yang_2021_ICCV} to parse each frame into a compact set of object tokens, serving as compressed representations for each frame. Thereafter, we feed these object tokens into object-time interaction layers to model the object state changes, which vividly depict the temporal dynamics in the video. 
Finally, we apply a linear head to the latent features representing these object state changes to produce the video-level prediction for action recognition.
Unlike the existing works on object-centric representations, our method does not rely on object annotations~\cite{herzig2022object,wang2018videos,zhang2022object}, extra detectors~\cite{zadaianchuk2022unsupervised} or additional modalities~\cite{zhang2024object}.
Instead, we develop two simple object-level losses to assist in distilling effective object-centric representations from image foundation models and establishing meaningful object state changes.

In this way, our method effectively mitigates the limitations of the previous work. Firstly, our method exhibits a notable enhancement in memory efficiency compared to conventional approaches. 
By exclusively finetuning the additional parameters on top of the image pre-trained model, we successfully reduce the memory required for gradient backpropagation. Additionally, compressing the video frames into object tokens significantly diminishes the computation requirements for temporal modeling, making our method more accessible and scalable.
Secondly, by utilizing object-centric reasoning in the form of learnable object tokens and establishing the temporal state changes of individual objects, we inject strong inductive bias into the learning process that goes beyond basic temporal modules like attention layers. We demonstrate that our model can localize the specific objects in the video as a side-product, providing an insightful interpretation of the image-to-video adaptation.

To summarize, our contributions are as follows. (1) We propose an efficient object-centric method to adapt image pre-trained models to the video domain. By parsing video frames into object tokens and establishing object state changes across time, we enhance temporal perception with reduced computation redundancy. (2) We achieve state-of-the-art performance on action recognition, presenting higher efficiency than existing efficient tuning counterparts. (3) Our method effectively discovers different objects in videos without object-level annotations and achieves robust zero-shot video object segmentation.

\section{Related Work}
\noindent\textbf{Video action recognition} is a fundamental problem in computer vision. Compared to image tasks, the extra temporal relationships between different frames are crucial for video understanding. To this end, prevalent methods build on existing image models~\cite{he2016deep,dosovitskiy2020image,liu2021swin}, and make architectural modifications to enable temporal modeling~\cite{carreira2017quo,liu2022video,qian2021enhancing,arnab2021vivit,lin2020space,ding2021motion,feichtenhofer2019slowfast,ding2022dual,qian2024streaming,tran2015learning,qian2022static,ding2023prune,tran2018closer}. These works either fully finetune the whole model~\cite{bertasius2021space,arnab2021vivit,carreira2017quo,feichtenhofer2019slowfast} or directly train on large-scale video data from scratch~\cite{tong2022videomae,qian2021spatiotemporal}, requiring huge computation cost. More recently, some works propose to recognize actions from the perspective of state changes~\cite{wang2016actions,peh2024learning,souvcek2022look,alayrac2024multi,nagarajan2018attributes}. This novel idea regards actions as transformations that make changes to the objects and environments. However, most of these works directly model the state changes on the representations of the entire scene, where different semantics and objects are highly entangled. It is non-trivial to accurately perceive the state changes of different objects. In this work, we explicitly decouple different objects from the video scene, and respectively establish the state changes on individual objects to assist action recognition.

\noindent\textbf{Object-centric video representation learning} aims at establishing robust representations for different objects to assist video understanding. To achieve this goal, a series of works borrow the idea from object discovery and employ iterative slot attention~\cite{locatello2020object} to decompose different objects~\cite{kipf2021conditional,kabra2021simone,crawford2020exploiting,besbinar2021self,ding2022motion,Yang_2021_ICCV,qian2023semantics,aydemir2023self,ding2023betrayed}. However, they are specifically trained for segmentation and can only deal with low-level segmentation tasks and lack the ability for high-level understanding. Another line of works uses the object-related features as a complement to the original video representations to enhance object awareness~\cite{herzig2022object,zhang2019structured,materzynska2020something,girdhar2019video,zhang2022object}. These methods achieve promising results on multiple action benchmarks~\cite{gu2018ava,goyal2017something,materzynska2020something}, but still have two major limitations. First, they depend on bounding box annotations~\cite{herzig2022object,wang2018videos,zhang2022object}, additional detectors or modalities~\cite{zadaianchuk2022unsupervised,zhang2024object} to identify different objects, restricting the scalability to unseen data. Second, the identified objects provide plentiful cues for video analysis, yet these methods still fuse the disentangled object representations with the original video features, imposing limitations on efficiency. In contrast, our process is free of object annotations or handcrafted priors but only adopts simple contrastive loss to distill general object knowledge from image foundation models and decouple object components in video frames. Then we directly model temporal state changes on compact object representations, enabling more intuitive and efficient video understanding.

\noindent\textbf{Efficient tuning} becomes a tending direction with the development of large foundation models~\cite{radford2021learning,zhou2021ibot,devlin2018bert,brown2020language}. It is proposed in natural language processing to reduce the trainable parameters and improve efficiency when transferring pre-trained models to downstream tasks~\cite{hu2021lora,houlsby2019parameter,lester2021power,zaken2021bitfit}. Common strategies include inserting task-specific adapters into Transformer encoders~\cite{pfeiffer2020adapterhub,pfeiffer2020adapterfusion,houlsby2019parameter}, attaching prompt tokens to the input~\cite{qin2021learning,shin2020autoprompt,liu2021p,li2021prefix} and learning low-rank approximations~\cite{hu2021lora}. Recently, the efficient tuning is increasingly explored in computer vision~\cite{jia2022visual,bahng2022visual,chen2022adaptformer,sung2022vl,zhou2022conditional,zhou2022learning,feng2022promptdet,liang2022open}. By freezing the image pre-trained model, \cite{sung2022vl,chen2022adaptformer} tune adapters with learnable parameters for efficient adaptation. \cite{jia2022visual,bahng2022visual,feng2022promptdet,liang2022open} develop prompt tuning to adapt the frozen image model to various image tasks. Later, \cite{ju2022prompting,lin2022frozen,ni2022expanding,yang2023aim,pan2022st,wang2021actionclip,Park_2023_CVPR, liu2023revisiting,Qing_2023_ICCV} take one step further to adapt image models to videos with temporal dynamics. In order to equip the model with temporal perception ability, \cite{yang2023aim,pan2022st,liu2023revisiting} insert spatio-temporal modules into Transformer encoder blocks, but the tunable parameters in shallow Transformer layers require huge GPU memory in gradient backpropagation.  \cite{Qing_2023_ICCV,Park_2023_CVPR} require two pathways to process spatial and temporal information. \cite{lin2022frozen,ju2022prompting} attach spatio-temporal attention or motion decoders to the frame-wise features, but there exists high redundancy in attention computation~\cite{tong2022videomae,bertasius2021space}.
% \cite{Park_2023_CVPR,Qing_2023_ICCV} introduce a new temporal path to process the temporal dynamics. 
Contrarily, our method retains the backbone architecture without modifications, distills object knowledge from image foundation models, and captures temporal state changes of discovered objects for video downstream tasks.

\begin{figure*}
    \centering
    \includegraphics[width=\linewidth]{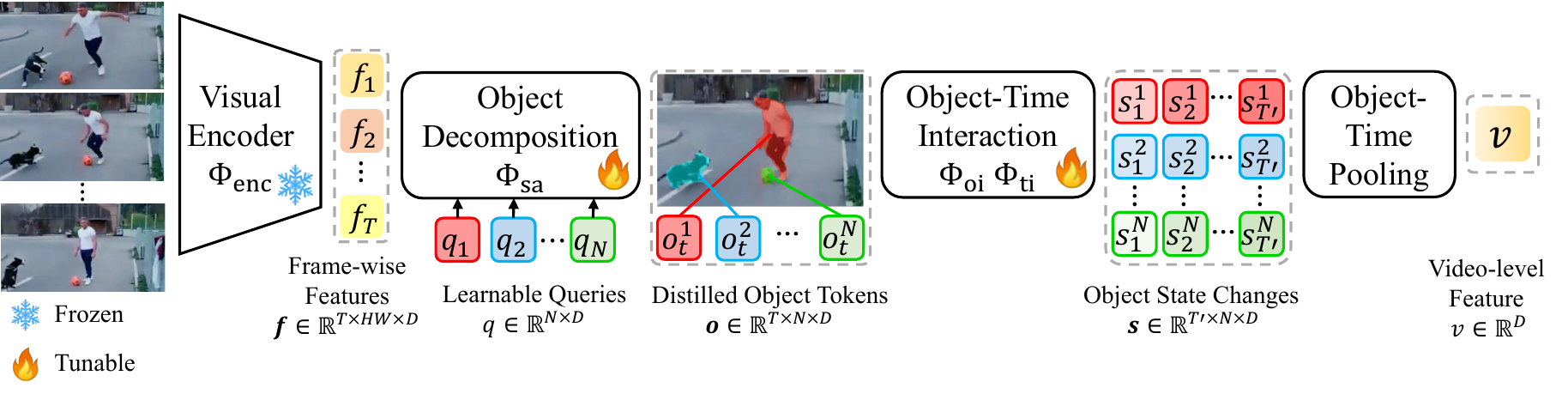}
    \caption{An overview of our object-centric image-to-video adaptation framework. We use a frozen image pre-trained model to extract frame-wise features and pass them through a lightweight temporal fusion block. Then we employ slot attention with learnable queries to decompose each frame into a compact set of object tokens. Thereafter, we develop object-time interaction layers to establish inter-object interactions and build temporal state changes of individual objects, which are then pooled into the video-level feature for action recognition.}
    \label{fig:arch}
\end{figure*}
\section{Method}
The overall architecture of our object-centric image-to-video adaptation is shown in Fig.~\ref{fig:arch}. We use a frozen pre-trained Vision Transformer~\cite{dosovitskiy2020image} as the backbone to extract frame-wise features from the video clip. Then we apply slot attention with learnable query tokens to identify different object components and pass the object tokens through object-time interaction layers to model the temporal state changes of individual objects. The obtained object state change vectors encode rich semantics and essential temporal dynamics in the given video, and are subsequently pooled into the final video-level feature for downstream tasks.

\subsection{Object Decomposition}
To start with, given a video clip $\bm{x}=\{x_1,x_2,\cdots,x_T\}$ consisting of $T$ frames, we use a frozen image pre-trained Vision Transformer ${\rm \Phi}_{\text{enc}}$ as the backbone to extract frame-wise features:
\begin{align}
    p_t, f_t = {\rm \Phi}_{\text{enc}}(x_t),\quad p_t\in\mathbb{R}^D,\quad f_t\in\mathbb{R}^{HW\times D},
\end{align}
where $p_t$ and $f_t$ denote the output \texttt{cls} token and feature map from the last Transformer block. $H,W,D$ respectively denote height, width, and channel dimension.
These image pre-trained features encode rich but highly-entangled semantics. And the next step is to parse these frame features into different object components which are crucial for video perception and understanding~\cite{zhang2022object,girdhar2019video,herzig2022object,wang2018videos}.

% Thereafter, we introduce a lightweight temporal fusion head, 
% ${\rm \Phi}_{\text{temp}}$, implemented as a temporal convolution layer with a kernel size of 3, to effectively leverage the temporal relationships between adjacent frames.

% To take advantage of the temporal relationship between adjacent frames, we introduce a lightweight temporal fusion head, denoted as ${\rm \Phi}_{\text{temp}}$. In practice, we implement the temporal fusion head ${\rm \Phi}_{\text{temp}}$ as one temporal convolution layer with a kernel size of 3.
% To utilize the temporal dependency between adjacent frames, a lightweight temporal fusion head ${\rm \Phi}_{\text{temp}}$ is introduced, via one temporal convolution layer with kernel size three in default:
% Formally, we obtain the $\Tilde{\bm{f}}\in\mathbb{R}^{T\times HW\times D}$ as follows:
% \begin{align}
%     \Tilde{\bm{f}}=\{\Tilde{f}_t\}_{t=1}^{T} = {\rm \Phi}_{\text{temp}}([f_1\oplus f_2\oplus\cdots\oplus f_T]),
% \end{align}
% where $[\oplus]$ indicates concatenation along time dimension.

% Till this point, with the fused features $\Tilde{\bm{f}}$, the next step is to identify different object components in each frame. 

Inspired by recent works on unsupervised object discovery and segmentation~\cite{locatello2020object,xu2022groupvit,xie2022segmenting}, we develop a variant of slot attention on top of the feature map $f_t$ for object decomposition in each frame. To be specific, instead of random sampling from a prior distribution, we use $N$ learnable query tokens $q\in\mathbb{R}^{N\times D}$ to initialize the slot vectors $S\in\mathbb{R}^{N\times D}$. The intuition of this design is to encourage each query token to capture specific semantics or concepts~\cite{xu2022groupvit,jia2022unsupervised} and produce temporally coherent object decompositions, which provide essential inductive bias for temporal reasoning in later stages. Then at each slot attention iteration, following~\cite{locatello2020object}, we use three learnable linear transformations to project the slot vectors $S$ and frame features $f_t$ into \texttt{query}, \texttt{key} and \texttt{value}, i.e., $Q\in\mathbb{R}^{N\times D}$, $K\in\mathbb{R}^{HW\times D}$ and $V\in\mathbb{R}^{HW\times D}$. After that, we compute the attention matrix $\Tilde{A}\in\mathbb{R}^{N\times HW}$ with softmax normalization along the slot dimension as follows:
\begin{align}
    \Tilde{A}_{i,j} := \frac{e^{A_{i,j}}}{\sum_{l=1}^N e^{A_{l,j}}}, \quad \text{where} \quad A := \frac{1}{\sqrt{D}}QK^T.
    \label{mask}
\end{align}
We use the weighted mean to aggregate the \texttt{value} and pass them through Gated Recurrent Unit (GRU) to update the slot vectors:
\begin{align}
    S := \text{GRU}(\texttt{inputs}=MV, \texttt{states}=S),\quad
    \text{where} \quad M_{i,j} := \frac{\Tilde{A}_{i,j}}{\sum_{l=1}^{HW}\Tilde{A}_{i,l}}.
\end{align}
We iterate the routing process three times, and take the final slot attention weights as potential object segmentation masks. The slot vectors from the final iteration are treated as the object tokens that distill object knowledge from the pre-trained image features. Note that we perform slot attention ${\rm \Phi}_{\text{sa}}$ on each frame in parallel, and denote the object tokens of each time stamp as:
\begin{align}
    \bm{o} = \{o_1\oplus o_2\oplus ...\oplus o_T\}\in\mathbb{R}^{T\times N\times D},
\end{align}
where $o_t = {\rm \Phi}_{\text{sa}}(q,\Tilde{f}_t)\in\mathbb{R}^{N\times D}$, and $\oplus$ denotes concatenation operation along time dimension.
Since this attention mechanism introduces competition among slots, the obtained object tokens $o_t$ are expected to take over distinct semantic parts and separate each frame into $N$ different components.

Surprisingly, we find that the distilled tokens and corresponding attention matrices perform remarkably well in \emph{zero-shot} video object segmentation tasks. They are even comparable to some pixel-level supervised counterparts, as demonstrated in Section~\ref{vos_exp}. This result verifies that our method effectively extracts robust object-centric features for temporal reasoning in the later stage. Note that since videos often contain partial object semantics, there tend to be some slots that carry background information and interact with the foreground object tokens for comprehensive video analysis.
% and in return, this temporal reasoning bolsters the process of object decomposition.

\subsection{Object-Time Interaction}
After obtaining the object tokens that reveal potential object semantics, we use them as compressed representations of video frames and further model the temporal dynamics from an object-centric perspective~\cite{zhang2019structured,herzig2022object}. Our intuition is that in continuous frames, the object tokens, $o_t^n\in\mathbb{R}^D, t=1,2,...,T$, with the same index $n$ tend to represent the same objects across time as illustrated in Fig.~\ref{teaser}. Hence, it is feasible to temporally track each object and conclude the object state changes based on the learned object tokens, thereby assisting the process of video understanding.

% To achieve this goal, we take inspiration from TimeSformer~\cite{bertasius2021space} and develop divided object-time attention for object-time interaction. The intuition is that in continuous video frames, the object tokens, $o_t^n\in\mathbb{R}^D, t=1,2,...,T$, with the same index $n$ tend to represent the same or similar objects across time as illustrated in Fig.~\ref{teaser}. Hence, it is feasible to alternatively compute attention on object and time dimension to respectively establish multi-object interactions and object-wise temporal dynamics.

% After obtaining the object tokens that reveal potential object semantics in each frame, we further model temporal dynamics alongside multi-object interactions~\cite{zhang2019structured,herzig2022object}. Drawing inspiration from TimeSformer~\cite{bertasius2021space}, we develop a divided object-time attention for this interaction. The intuition is that the object tokens, $o_t^n\in\mathbb{R}^D, t=1,2,...,T$, with index $n$ typically represent similar objects over time, as seen in Fig.~\ref{teaser}. It's thus effective to calculate attention on object and time dimensions for multi-object interactions and object-wise temporal dynamics. 

To achieve this goal, we first apply self-attention to the object tokens within the same frame to introduce inter-object interactions at each time stamp:
\begin{align}
    \Tilde{o}_t = {\rm \Phi_{oi}}(o_t)\in\mathbb{R}^{N\times D},
\end{align}
where $\rm \Phi_{oi}$ is the object interaction module instantiated by a standard Transformer encoder layer consisting of linear projections, self-attention calculation and a feed forward network. The resulting vector $\Tilde{o}_t^n$ not only encodes the attribute of object $n$ but also captures its relationships with contextual objects, serving as a representation for the latent state of object $n$ at time stamp $t$.

Considering that the temporal state changes are a concrete manifestation of the effects of actions, they serve as an important cue for analyzing the events occurring in a video and assisting video understanding~\cite{wang2016actions,souvcek2022look,alayrac2024multi,peh2024learning}. Hence, based on the vectors representing latent object states, we take one step further to look into multiple time stamps and explicitly establish the state changes of each object. Without loss of generality, we take the object with index $n$ for illustration. Given the object states at different time stamps, we define a temporal interval $\delta$ to sample the initial and final states, i.e., $\Tilde{o}_t^n$ as the initial state and $\Tilde{o}_{t+\delta}^n$ as the final state. We concatenate them along channel dimension and pass them through the temporal interaction layer $\rm \Phi_{ti}$ to model the object state change between time $t$ and $t+\delta$:
\begin{align}
    s_t^n = {\rm \Phi_{ti}}\left(\left[\Tilde{o}_t^n\oplus \Tilde{o}_{t+\delta}^n\right]\right)\in\mathbb{R}^D,\quad t=\{1,2,...,T-\delta\},
\end{align}
where $\oplus$ denotes channel-wise concatenation.
We instantiate $\rm \Phi_{ti}$ as a multi-layer perceptron of channel dimension $2D\text{-}D\text{-}D$. In this way, for each object, we obtain $T'=T-\delta$ state change vectors. The state change matrix covering all objects and time stamps can be denoted as $\bm{s}\in\mathbb{R}^{T'\times N\times D}$.

Since the state change is a continuous process within a video sequence, it is feasible to conclude the overall state change of each object throughout the entire video by simple temporal average pooling on the established matrix $\bm{s}$. Thereafter, we further apply average pooling on object dimension to aggregate the state change information of all objects as the final video-level representation, $v\in\mathbb{R}^D$, for downstream action recognition.

\noindent\textbf{Discussion.} Unlike the existing works that use object features as complements to original video features~\cite{herzig2022object,zhang2019structured,materzynska2020something,girdhar2019video,zhang2022object} or model state changes on the highly entangled scene representations~\cite{wang2016actions,peh2024learning,souvcek2022look,alayrac2024multi}, we perform temporal reasoning solely on distilled object tokens. Our object-centric state change modeling reduces redundancy, mitigates ambiguity and results in more fine-grained cues for video understanding.
And comparing with the prevalent image-to-video adaptation methods, they introduce dense temporal fusion operations, e.g., temporal convolutions~\cite{pan2022st,lin2022frozen,liu2023revisiting} and attention~\cite{yang2023aim,Qing_2023_ICCV,Park_2023_CVPR,lin2022frozen,liu2023revisiting}, to mix multi-frame features for action recognition. In contrast, we leverage the object state changes to perceive temporal dynamics, which closely resemble human logic and provide an effective inductive bias to the model. To be specific, it does not require dense fusion or traversal of all frames. Instead, we respectively model state changes on separated objects across sparsely sampled frames, enabling more efficient scene parsing and temporal perception as verified in Section~\ref{ablation_exp}.

\subsection{Training}
In training, we freeze the parameters in the image pre-trained model ${\rm \Phi}_{\text{enc}}$, and tune parameters that are responsible for object decomposition and temporal perception, i.e., ${\rm \Phi}_{\text{temp}}$, ${\rm \Phi}_{\text{oi}}$ and ${\rm \Phi}_{\text{ti}}$. Besides the standard cross-entropy loss for action classification, we introduce two object-level losses to further facilitate object decomposition and temporal state change modeling.

\noindent\textbf{Object Distillation Loss.}
Considering that the image pre-trained foundation models tend to encode rich semantics in the given scene~\cite{radford2021learning,caron2021emerging,oquab2023dinov2}, the output \texttt{cls} token $p_t$ is expected to encompass the majority of the semantic components in frame $t$. Hence, we use a contrastive grounding loss~\cite{zareian2021open,gupta2020contrastive,ghiasi2022scaling} to align the learned object tokens $o_t$ and the \texttt{cls} token $p_t$, encouraging $o_t$ to cover all objects in frame $t$. Particularly, we first define the correspondence score between the $o_t$ and $p_t$ as:
\begin{align}
    c(o_t,p_t) = \sum_{n=1}^N\frac{e^{\left\langle o_t^n,p_t\right\rangle/\tau}}{\sum_{l=1}^N e^{\left\langle o_t^l,p_t\right\rangle/\tau}}\left\langle o_t^n,p_t\right\rangle,
\end{align}
where $\left\langle o_t^n,p_t\right\rangle=\frac{o_t^n p_t}{\norm{o_t^n}\norm{p_t}}$ denotes cosine similarity calculation, $\tau$ is the temperature hyper-parameter. This correspondence score adaptively aggregates each object component and avoids penalizing unrelated object semantics. Thereafter, we randomly sample \texttt{cls} token vectors $p'$ from other videos within the mini-batch to formulate a negative sample pool $\mathcal{N}$ and apply the contrastive loss to maximize the agreement between the aligned object and \texttt{cls} token pairs:
\begin{align}
    \mathcal{L}_{obj} = -\sum_{t=1}^T\log\frac{e^{c(o_t,p_t)/\tau}}{e^{c(o_t,p_t)/\tau}+\sum_{p'\sim\mathcal{N}}e^{c(o_t,p')/\tau}}.
\end{align}
This self-supervised alignment objective encourages the object tokens to cover diverse semantics and distill object knowledge from the image pre-trained model.

\noindent\textbf{Temporal Reasoning Loss.}
In our architecture, we analyze temporal dynamics from the perspective of object state changes. Therefore, it is crucial to formulate a learning objective that guides the temporal interaction module $\rm \Phi_{ti}$ to accurately capture the state changes between different time stamps. To achieve this goal, our intuition is that the same objects undergoing the same type of actions should possess consistent state changes. Conversely, different actions will result in distinct state changes for the same object, and different objects will exhibit diverse state changes under the same action. Based on this, we develop a margin loss to guide object state change modeling. In detail, given $s_t^n$ as query, we define the videos within the mini-batch that have the same action labels with the current video to form positive pool $\mathcal{P}$, and those of different action labels as negative pool $\mathcal{N}$. We sample the state change vectors of the same object and the same action, i.e., $\{\hat{s}_t^n|\hat{s}\in\mathcal{P}\}$, as positive pairs, which should be aligned with $s_t^n$. As for the negatives, we respectively sample the state change vectors of the same object but different actions, i.e., $\{s'_t{}^n|s'\in\mathcal{N}\}$, and those from the same video but of different objects, i.e., $\{s_t^m,m\neq n\}$, as \textit{hard negatives} for discrimination. The margin loss with a margin hyper-parameter $\lambda$ is formulated as:
\begin{align}
\begin{split}
    \mathcal{L}_{temp} = &\sum_{t=1}^{T'}\sum_{n=1}^N\{\sum_{\hat{s}\in\mathcal{P}}\norm{s_t^n-\hat{s}_t^n}_2\\
    &+\sum_{s'\in\mathcal{N}}\max(\lambda-\norm{s_t^n-s'_t{}^n}_2, 0)
    +\sum_{m=1\atop m\neq n}^N\max(\lambda-\norm{s_t^n-s_t^m}_2, 0)\},
\end{split}
\end{align}
where $\norm{\cdot}_2$ denotes $\mathcal{L}_2$ distance. This term enforces the model to discriminate diverse object states and capture unique dynamic effects of each action category.

\noindent\textbf{Overall Objectives.}
Besides the above object-level losses, for action recognition tasks, we apply a linear classification head on the video-level representation $v$ with a supervised classification loss $\mathcal{L}_{cls}$ in the form of cross-entropy. In this way, we formulate the overall loss function as:
\begin{align}
    \mathcal{L} = \mathcal{L}_{obj} + \mathcal{L}_{temp} + \mathcal{L}_{cls}.
\end{align}
% We set $\lambda=1$ when adapting the model for action recognition, and set $\lambda=0$ to guide the architecture to transfer object knowledge from image pre-trained models to video domain in a fully self-supervised manner. We empirically find that the self-supervised training is sufficient to enable the model to identify objects and formulate object-centric representations for video analysis.

\section{Experiments}

% \subsection{Dataset}
% We train and evaluate three popular action recognition dataset, Kinetics-400 (K-400)~\cite{kay2017kinetics}, Something-Something-v2 (SSv2)~\cite{goyal2017something} and Epic-Kitchens-100 (EK100)~\cite{damen2020rescaling}. The Kinetics-400 dataset, with around 240k training and 20k validation video clips, covers 400 distinct action categories. Each clip has a duration of approximately 10 seconds. The Something-Something V2 dataset encompasses 174 categories and requires strong temporal modeling due to the consistent background across various action classes. This dataset includes approximately 170k training videos and 25k validation videos. The Epic-Kitchens-100 is an egocentric dataset annotated with verbs and nouns, with 495/138/67 videos for train/validation/test. Each video contains long-term object interactions and challenging camera movements.
% Besides, we also evaluate the ability of zero-shot video object segmentation on DAVIS-2016~\cite{perazzi2016benchmark}, SegTrack-v2~\cite{li2013video}, FMBS-59~\cite{ochs2013segmentation} and more challenging multiple object segmentation on DAVIS-2017-UVOS~\cite{caelles20192019}. 
% We respectively report the mean per frame intersection over union (IoU) and $\mathcal{J}\&\mathcal{F}$ on single and multiple object discovery benchmarks.

\subsection{Implementation Details}
We train and evaluate on three popular action recognition datasets: Kinetics-400 (K-400)~\cite{kay2017kinetics}, Something-Something-v2 (SSv2)~\cite{goyal2017something}, and Epic-Kitchens-100 (EK100)~\cite{damen2020rescaling}. Additionally, we assess zero-shot video object segmentation on challenging multiple object segmentation dataset, DAVIS-2017-UVOS~\cite{caelles20192019}. 
We use the frozen CLIP~\cite{radford2021learning} pre-trained Vision Transformer~\cite{dosovitskiy2020image} as ${\rm \Phi}_{\text{enc}}$ for frame-wise feature extraction. In training, we crop each frame into $224\times 224$, set the number of query tokens $N=8$, and the time interval for object state change $\delta=T/4$ in default. We use AdamW optimizer~\cite{loshchilov2018decoupled} with an initial learning rate $5\times 10^{-4}$ and mini-batch size $512$ to update the tunable parameters. %For more comprehensive implementation details, please refer to the Supplementary Materials.
% For unsupervised video object segmentation, we employ the slot attention weight of each object token as a candidate object mask, then report IoU and $\mathcal{J}\&\mathcal{F}$ scores~\cite{caelles20192019}.

\begin{table*}[t]
    \centering
    \small
    \caption{Results on Kinetics-400. We report the pre-train model initialization, inference GFLOPs, the number of tunable parameters (M), Top-1 and Top-5 accuracy. The `Views' indicates Frame Number $\times$ Temporal Clips $\times$ Spatial Crops.}
    \begin{tabular}{lcccccc}
    \toprule
        Method & Pre-train & Views & GFLOPs & Tunable & Top-1 & Top-5 \\
        \midrule
        \textit{Full finetuning} \\
        % MViT-B~\cite{fan2021multiscale} & - & $64\times 3\times 3$ & 4095 & 37 & 81.2 & 95.1 \\
        % UniFormer-B~\cite{li2022uniformer} & IN-1K & $32\times 4\times 3$ & 3108 & 50 & 83.0 & 95.4 \\
        TimeSformer-L~\cite{bertasius2021space} & IN-21K & $64\times 1\times 3$ & 7140 & 121 & 80.7 & 94.7 \\
        % ViViT-L/16$\times$2 FE~\cite{arnab2021vivit} & IN-21K & $32\times 1\times 1$ & 3980 & 311 & 80.6 & 92.7 \\
        VideoSwin-L~\cite{liu2022video} & IN-21K & $32\times 4\times 3$ & 7248 & 197 & 83.1 & 95.9 \\
        % MViTv2-L (312$\uparrow$)~\cite{li2022mvitv2} & IN-21K & $32\times 3\times 5$ & 42420 & 218 & 86.1 & 97.0 \\
        MTV-L~\cite{yan2022multiview} & JFT & $32\times 4\times 3$ & 18050 & 876 & 84.3 & 96.3 \\
        % TokenLearner-L/10~\cite{ryoo2021tokenlearner} & JFT & $64\times 4\times 3$ & 48192 & 450 & 85.4 & 96.3 \\
        PromptCLIP A7~\cite{ju2022prompting} & CLIP & $16\times 5\times 1$ & - & - & 76.8 & 93.5 \\
        ActionCLIP~\cite{wang2021actionclip} & CLIP & $32\times 10\times 3$ & 16890 & 142 & 83.8 & 97.1 \\
        X-CLIP-L/14~\cite{ni2022expanding} & CLIP & $8\times 4\times 3$ & 7890 & 420 & 87.1 & 97.6 \\
        \midrule
        \textit{Efficient tuning} \\
        EVL ViT-L/14~\cite{lin2022frozen} & CLIP & $32\times 3\times 1$ & 8088 & 59 & 87.3 & - \\
        ST-Adapter ViT-B/16~\cite{pan2022st} & CLIP & $32\times 3\times 1$ & 1821 & 7.2 & 82.7 & 96.2 \\
        STAN-self-B/16~\cite{liu2023revisiting} & CLIP & $16\times 3\times 1$ & 1187 & - & 84.9 & 96.8 \\
        AIM ViT-B/16~\cite{yang2023aim} & CLIP & $32\times 3\times 1$ & 2428 & 11 & 84.7 & 96.7 \\
        % AIM ViT-L/14~\cite{yang2023aim} & CLIP & $32\times 3\times 1$ & 11208 & 38 & 87.5 & 97.7 \\
        DualPath ViT-B/16~\cite{Park_2023_CVPR} & CLIP & $32\times 3\times 1$ & 710 & 10 & 85.4 & 97.1 \\
        DiST ViT-B/16~\cite{Qing_2023_ICCV} & CLIP & $32\times 3\times 1$ & 1950 & - & 85.0 & 97.0 \\
        DiST ViT-L/14~\cite{Qing_2023_ICCV} & CLIP & $32\times 3\times 1$ & 8490 & - & 88.0 & 97.9 \\
        \midrule
        % Ours ViT-S/16 & DINO & $8\times 3\times 1$ & 107 & 1.7 & 80.4 & 93.1 \\
        \textbf{Ours ViT-B/16} & CLIP & $8\times 3\times 1$ & 281 & 6.5 & 85.8 & 97.3 \\
        \textbf{Ours ViT-B/16} & CLIP & $16\times 3\times 1$ & 562 & 6.5 & 86.1 & 97.5 \\
        \textbf{Ours ViT-B/16} & CLIP & $32\times 3\times 1$ & 1123 & 6.5 & 86.2 & 97.5 \\
        \textbf{Ours ViT-L/14} & CLIP & $32\times 3\times 1$ & 5068 & 21 & \textbf{88.5} & \textbf{98.0} \\
        % Ours ViT-S/14 & DINOv2 & $8\times 3\times 1$ & 139 & 1.7 & 81.1 & 92.9 \\
        % Ours ViT-B/14 & DINOv2 & $8\times 3\times 1$ & 551 & 6.7 & 83.7 & 95.8 \\
        % Ours ViT-B/14 & DINOv2 & $32\times 3\times 1$ & 2205 & 6.7 & 84.9 & 96.8 \\
        % Ours ViT-L/14 & DINOv2 & $32\times 3\times 1$ & 7642 & 21 & 88.1 & \textbf{97.9} \\
        \bottomrule
    \end{tabular}
    \label{k400}
\end{table*}

\subsection{Video Action Recognition}
In this section, we present the comparisons between our method and recent state-of-the-art on K-400, SSv2 and EK100 in terms of the number of tunable parameters, inference GFLOPs and action recognition accuracy. 

\noindent\textbf{K-400.} We first show the results on K-400 in Table~\ref{k400} and conclude three observations. (1) Compared to fully finetuned models, our method achieves superior performance with significantly fewer tunable parameters. For example, our method of ViT-L/14 with only 21M tunable parameters outperforms all the fully finetuned approaches that require over 100M tunable parameters. (2) Our method with only a visual encoder achieves better results than the multi-modal methods, ActionCLIP~\cite{wang2021actionclip} and X-CLIP~\cite{ni2022expanding}, which employ an extra text branch to guide the finetuning process.
(3) Among the efficient tuning methods~\cite{lin2022frozen,yang2023aim,pan2022st,liu2023revisiting,Park_2023_CVPR,Qing_2023_ICCV}, EVL~\cite{lin2022frozen} attaches a motion decoder with multiple temporal convolution and attention layers for temporal modeling. AIM~\cite{yang2023aim}, ST-Adapter~\cite{pan2022st} and STAN~\cite{liu2023revisiting} insert temporal blocks into intermediate CLIP encoder layers. DualPath~\cite{Park_2023_CVPR} and DiST~\cite{Qing_2023_ICCV} introduce an additional temporal pathway to capture temporal dynamics. Our formulation models temporal relationships on decomposed objects in the form of state changes, which reduces computation redundancy but attains superior performance. For example, on ViT-B/16, our method with only an 8-frame input yields superior results to other approaches that employ a 32-frame input. On ViT-L/14, we reach the state-of-the-art results with 40\% fewer GFLOPs (5068 vs 8490) than DiST~\cite{Qing_2023_ICCV}.
% (3) Our method is applicable to various image pre-trained models, e.g., DINO~\cite{caron2021emerging}, CLIP~\cite{radford2021learning}, DINOv2~\cite{oquab2023dinov2}, demonstrating the feasibility of transferring object knowledge from image pre-trained models to assist video analysis.

\noindent\textbf{SSv2.} When extending to SSv2 which requires stronger temporal modeling, the observations are consistent. As depicted in Table~\ref{ssv2}, our object-centric formulation achieves state-of-the-art results and presents higher efficiency than both full finetuning and efficient tuning methods. In particular, our method with ViT-L/14 surpasses the advanced MViTv2-L~\cite{li2022mvitv2} and DiST~\cite{Qing_2023_ICCV} with significantly reduced GFLOPs (5068 vs 8484, 8490). Moreover, on ViT-B/16, our method with only 8 frames as input outperforms others using 32 frames. It is non-trivial to achieve this on such a motion-heavy dataset, which in return reveals the significance of our temporal modeling in the form of object state changes. On the one hand, it enables accurate temporal perception with sparser sampling. On the other hand, our temporal modeling on individual objects eliminates the need for dense temporal fusion, substantially reducing computational redundancy.
% It outperforms most of the comparing approaches with a 32-frame input, revealing that our object-centric formulation leads to more efficient temporal modeling. Moreover, our method with ViT-L/14 backbone obtains comparable performance to MViTv2-L~\cite{li2022mvitv2} and DiST~\cite{Qing_2023_ICCV} with significantly reduced GFLOPs (5154 vs 8484, 8490). This phenomenon reveals that the distilled object knowledge indeed reduces redundancy in temporal modeling, and demonstrates the potential of establishing temporal relationships on top of the decomposed objects.
% Our object-centric formulation outperforms all other methods with a frozen backbone with minimal computation cost. Despite that the fully finetuned MViTv2~\cite{li2022mvitv2} slightly exceeds our method, but \cite{li2022mvitv2} uses additional K-400 dataset to assist image-to-video adaptation and requires substantially more tunable parameters. This phenomenon reveals that the distilled object knowledge indeed reduces redundancy in temporal modeling, and demonstrates the potential of establishing temporal relationships on top of the decomposed objects.

\begin{table*}[t]
    \centering
    \small
    \caption{Results on Something-Something-v2. K-400$^\dagger$/K-600$^\dagger$ indicates the model is pre-trained on both IN-21K and K-400/K-600.}
    \begin{tabular}{lcccccc}
    \toprule
        Method & Pre-train & Views & GFLOPs & Tunable & Top-1 & Top-5 \\
        \midrule
        \textit{Full finetuning} \\
        TimeSformer-L~\cite{bertasius2021space} & IN-21K & $64\times 1\times 1$ & 7140 & 121 & 62.4 & - \\
        MTV-B~\cite{yan2022multiview} & IN-21K & $32\times 4\times 3$ & 4790 & 310 & 67.6 & 90.4 \\
        % MViT-B~\cite{fan2021multiscale} & K-400 & $32\times 1\times 3$ & 510 & 37 & 67.1 & 90.8 \\
        % MViTv2-B~\cite{li2022mvitv2} & K-400 & $40\times 1\times 3$ & 675 & 51 & 70.5 & 92.7 \\
        ViViT-L/16$\times$2~\cite{arnab2021vivit} & K-400$^\dagger$ & $16\times 4\times 3$ & 11892 & 311 & 65.4 & 89.8 \\
        VideoSwin-B~\cite{liu2022video} & K-400$^\dagger$ & $32\times 1\times 1$ & 963 & 89 & 69.6 & 92.7 \\
        % Omnivore~\cite{girdhar2022omnivore} & K-400$^\dagger$ & $32\times 1\times 3$ & - & - & 71.4 & 93.5 \\
        MViTv2-L (312$\uparrow$)~\cite{li2022mvitv2} & K-400$^\dagger$ & $32\times 1\times 3$ & 8484 & 213 & 73.3 & 94.1 \\
        % UniFormer-B~\cite{li2022uniformer} & K-600$^\dagger$ & $32\times 1\times 3$ & 777 & 50 & 71.2 & 92.8 \\
        \midrule
        \textit{Efficient tuning} \\
        EVL ViT-L/14~\cite{lin2022frozen} & CLIP & $32\times 1\times 3$ & 9641 & 175 & 66.7 & - \\
        ST-Adapter ViT-B/16~\cite{pan2022st} & CLIP & $32\times 3\times 1$ & 1955 & 7.2 & 69.5 & 92.6 \\
        STAN-self-B/16~\cite{liu2023revisiting} & CLIP & $16\times 3\times 1$ & 1376 & - & 69.5 & 92.7 \\
        AIM ViT-B/16~\cite{yang2023aim} & CLIP & $32\times 1\times 3$ & 2496 & 14 & 69.1 & 92.2 \\
        % AIM ViT-L/14~\cite{yang2023aim} & CLIP & $32\times 1\times 3$ & 11508 & 50 & 70.6 & 92.7 \\
        DualPath ViT-B/16~\cite{Park_2023_CVPR} & CLIP & $32\times 1\times 3$ & 716 & 13 & 70.3 & 92.9 \\
        DiST ViT-B/16~\cite{Qing_2023_ICCV} & CLIP & $32\times 3\times 1$ & 1950 & - & 70.9 & 92.1 \\
        DiST ViT-L/14~\cite{Qing_2023_ICCV} & CLIP & $32\times 3\times 1$ & 8490 & - & 73.1 & 93.2 \\
        \midrule
        % Ours ViT-S/16 & DINO & $8\times 1\times 3$ & 107 & 1.7 & 60.4 & 87.6 \\
        \textbf{Ours ViT-B/16} & CLIP & $8\times 1\times 3$ & 281 & 6.5 & 71.2 & 93.1 \\
        \textbf{Ours ViT-B/16} & CLIP & $16\times 1\times 3$ & 562 & 6.5 & 71.6 & 93.3 \\
        \textbf{Ours ViT-B/16} & CLIP & $32\times 1\times 3$ & 1123 & 6.5 & 71.8 & 93.4 \\
        \textbf{Ours ViT-L/14} & CLIP & $32\times 1\times 3$ & 5068 & 21 & \textbf{73.6} & \textbf{94.3} \\
        % Ours ViT-S/14 & DINOv2 & $8\times 1\times 3$ & 139 & 1.7 & 63.3 & 89.1 \\
        % Ours ViT-B/14 & DINOv2 & $8\times 1\times 3$ & 551 & 6.7 & 67.1 & 90.8 \\
        % Ours ViT-B/14 & DINOv2 & $32\times 1\times 3$ & 2205 & 6.7 & 70.1 & 92.6 \\
        % Ours ViT-L/14 & DINOv2 & $32\times 1\times 3$ & 7642 & 21 & 72.7 & 93.8 \\
        \bottomrule
    \end{tabular}
    \label{ssv2}
\end{table*}

\noindent\textbf{EK100.}
On challenging egocentric EK100 that requires long-term object interactions, our method showcases substantial advantages over existing efficient tunning counterparts~\cite{lin2022frozen,pan2022st,Qing_2023_ICCV} in Table~\ref{ek100}. The underlying reasons are two-fold. First, our method involves explicit object decomposition, which helps to capture the primary objects in egocentric videos and contributes to the significant improvements in nouns. Second, our temporal modeling on individual objects reduces the interference of camera motions and enforces the model to focus on the dynamics of each object, thus facilitating verb perception.

\subsection{Zero-shot Video Object Segmentation}
\label{vos_exp}
To verify whether our method understands videos in an object-centric way and interpret why the model works, we evaluate unsupervised video object segmentation in addition to action recognition. 
% We take the frozen DINO pre-trained ViT-S/16 as ${\rm \Phi}_{\text{enc}}$.
Particularly, we use the frozen ViT-B/16 backbone and adopt slot attention weights $\{\Tilde{A}_{i}\in\mathbb{R}^{HW}\}_{i=1}^{N}$ in Eq.~\ref{mask} as object masks candidate without bells and whistles. We report $\mathcal{J}\&\mathcal{F}$ on challenging multiple object discovery benchmark, DAVIS-2017-UVOS~\cite{caelles20192019}.
% use the frozen DINO pre-trained ViT-S/16 as ${\rm \Phi}_{\text{enc}}$ and train the architecture in a fully self-supervised manner, i.e., only $\mathcal{L}_{obj}$ and $\mathcal{L}_{temp}$. 
% Note that unlike previous studies that directly train on video segmentation benchmarks, our approach involves training on the K-400 or SSv2 datasets. 
Note that we directly apply the model trained on K-400 or SSv2 to video object segmentation in a \emph{zero-shot} manner.

\begin{table}[]
\begin{minipage}{0.55\linewidth}
    \centering
    \small
    \caption{Results on Epic-Kitchens-100 Verb, Noun and Action. All compared methods take an input of $8\times 1\times 3$.}
    \begin{tabular}{lccc}
    \toprule
        Method & Verb & Noun & Action \\
        \midrule
        EVL ViT-B/16~\cite{lin2022frozen} & 62.7 & 51.0 & 37.7 \\
        ST-Adapter ViT-B/16~\cite{pan2022st} & 67.7 & 55.0 & - \\
        DiST ViT-B/16~\cite{Qing_2023_ICCV} & 69.5 & 58.1 & 45.8 \\
        DiST ViT-L/14~\cite{Qing_2023_ICCV} & 70.7 & 61.6 & 48.9 \\
        \midrule
        \textbf{Ours ViT-B/16} & 71.1 & 60.9 & 49.0 \\
        \textbf{Ours ViT-L/14} & \textbf{73.2} & \textbf{64.5} & \textbf{51.7} \\
    \bottomrule
    \end{tabular}
    \label{ek100}
\end{minipage}
% \hspace{0.02\linewidth}
\begin{minipage}{0.44\linewidth}
    \centering
    \small
    \caption{Results on multiple object segmentation on DAVIS-2017-UVOS.}
    \begin{tabular}{lccc}
    \toprule
        Model & $\mathcal{J}\&\mathcal{F}$ & $\mathcal{J}$ & $\mathcal{F}$ \\
        \midrule
        % SAM~\cite{kirillov2023segment} & 12.3 & 10.2 & 14.4 \\
        % DINOSAUR~\cite{seitzer2022bridging} & 21.7 & 19.6 & 23.8 \\
        OCLR~\cite{xie2022segmenting} & 39.6 & 38.2 & 41.1 \\
        % SOLV~\cite{aydemir2023self} & - & 30.2 & - \\
        SMTC~\cite{qian2023semantics} & 40.5  & 36.4 & 44.6 \\
        % VideoCutLer~\cite{wang2023videocutler} & 43.6 & 41.7 & 45.5 \\
        BA~\cite{ding2023betrayed} & 43.9 & 39.2 & 48.6 \\
        \textcolor[rgb]{0.5, 0.5, 0.5}{RVOS}~\cite{ventura2019rvos} & \textcolor[rgb]{0.5, 0.5, 0.5}{41.2} & \textcolor[rgb]{0.5, 0.5, 0.5}{36.8} & \textcolor[rgb]{0.5, 0.5, 0.5}{45.7} \\
        \midrule
        \textbf{Ours SSv2} & 41.0 & 37.6 & 44.4 \\
        \textbf{Ours K-400} & \textbf{44.7} & \textbf{42.3} & \textbf{47.1} \\
        \textbf{Ours K-400 CRF} & \textbf{49.3} & \textbf{46.5} & \textbf{52.1} \\
        \bottomrule
    \end{tabular}
    \label{multiple}    
\end{minipage}
\end{table}

% \noindent\textbf{Single object segmentation.}
% We first compare the results on single object segmentation benchmarks without any post-processing in Table~\ref{single}. Most of the existing works resort to optical flow as input~\cite{Yang_2021_ICCV,yang2019unsupervised,xie2022segmenting,lian2023bootstrapping} or supervision~\cite{choudhury2022guess} to figure out moving pixels. \cite{liu2021emergence} exploits inter-frame warping to delineate object candidates. \cite{lian2023bootstrapping} introduces complicated appearance-motion synergy to enhance both static and dynamic object perception. Though not specifically designed for this task, our method with only RGB frames achieves comparable or superior results on three benchmarks. This promising zero-shot transfer performance demonstrates that our formulation manages to distill general object knowledge from image pre-trained models. Interestingly, the result trained on K-400 is significantly better than those on SSv2. We conjecture that it is because K-400 covers more diverse scenes and object semantics than SSv2. Thus, our model trained on K-400 presents a stronger zero-shot generalization ability to downstream video object segmentation.

\noindent\textbf{Quantitative Results.} We present the comparison with recent state-of-the-arts without object annotations~\cite{xie2022segmenting,qian2023semantics,wang2023videocutler} as well as a fully supervised baseline~\cite{ventura2019rvos} (denoted in \textcolor[rgb]{0.5, 0.5, 0.5}{grey}) in Table~\ref{multiple}. 
% Due to the lack of self-supervised counterparts that can deal with unsupervised multiple object discovery, we re-train DINOSAUR~\cite{seitzer2022bridging} on DAVIS-2017-UVOS, and run SAM~\cite{kirillov2023segment} inference without prompts for more comprehensive comparison. We observe that the recent advanced SAM performs comparatively poor in the absence of the prompts.
Among the compared works without mask annotations, OCLR~\cite{xie2022segmenting} takes optical flow as input to distinguish motion patterns of diverse objects. SMTC~\cite{qian2023semantics} designs two-stage slot attention to produce temporally consistent object segmentations.  %VideoCutLer~\cite{wang2023videocutler} synthesizes videos consisting of objects segmented from image pre-trained segmentor~\cite{cheng2022masked} and learns video segmentation from pseudo masks. 
BA~\cite{ding2023betrayed} also utilizes image foundation models like DION~\cite{caron2021emerging} to provide spatio-temporal correspondence and use clustering to predict the object segmentation masks.
On the one hand, since K-400 covers more diverse scenes and objects than SSv2, it leads to better generalization ability in this zero-shot segmentation scenario. On the other hand, our formulation trained on K-400 achieves superior results to these counterparts, even surpassing the simple supervised baseline RVOS~\cite{ventura2019rvos}. This demonstrates the feasibility of transferring object knowledge from image foundation models to the video domain, and our object-centric temporal reasoning further facilitates temporally coherent object decomposition. However, the gap to the state-of-the-art supervised method is still large because the mask annotations lead to much more accurate segmentation boundaries. A straightforward way is to apply Conditional Random Field (CRF)~\cite{lafferty2001conditional} for refinement, which leads to 4.6 points improvment on $\mathcal{J}\&\mathcal{F}$ score. It is worth further exploration in the future.

\noindent\textbf{Qualitative Results.} We also visualize some examples of the generated segmentation masks in Fig.~\ref{fig:seg}, where each column presents the frames within the same video and the same color denotes the masks generated by the same slot query. Generally, our model discriminates different object semantics in videos. Each distilled object token can capture specific semantics or concepts that generalize across diverse time stamps and scenes. For example, the token marked in \textcolor[rgb]{1, 0, 0}{red} focuses on humans, the \textcolor[rgb]{0, 0, 1}{blue} one emphasizes quadruped animals, and the \textcolor[rgb]{1, 0.7, 0}{yellow} one identifies two-wheeled vehicles. This temporally coherent object decomposition provides an effective inductive bias for subsequent temporal reasoning, leading to more efficient temporal modeling.

% In conclusion, our proposed method successfully learns object-centric representations and performs effective temporal reasoning based on the identified objects.

\subsection{Ablation Study}
\label{ablation_exp}
In this section, we use the CLIP ViT-B/16 as ${\rm \Phi}_{\text{enc}}$ with an input of 16 frames for all ablative experiments unless otherwise specified. More details and extra ablation studies are presented in Supplementary Materials.

\begin{figure}[]
    \begin{minipage}{0.55\linewidth}
        \centering
        \includegraphics[width=1.02\linewidth]{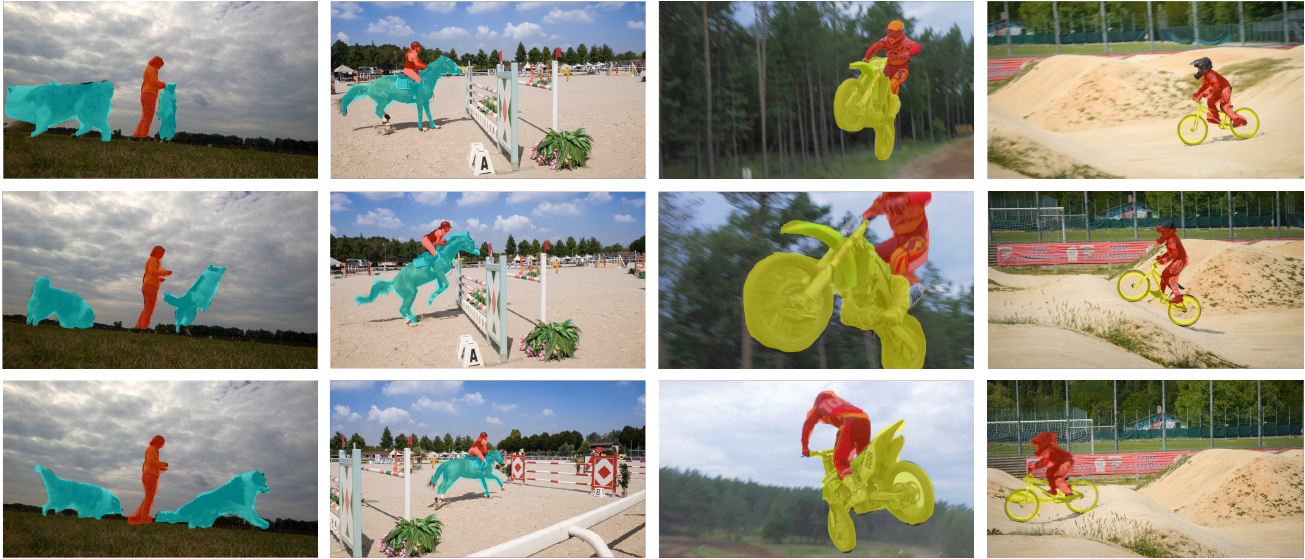}
        \caption{Visualization of object decomposition in the form of segmentation. Each column presents three frames in a video. The same color denotes the objects identified by the same object token.}
        \label{fig:seg}
    \end{minipage}
    \begin{minipage}{0.42\linewidth}
        \centering
        \includegraphics[width=0.93\linewidth]{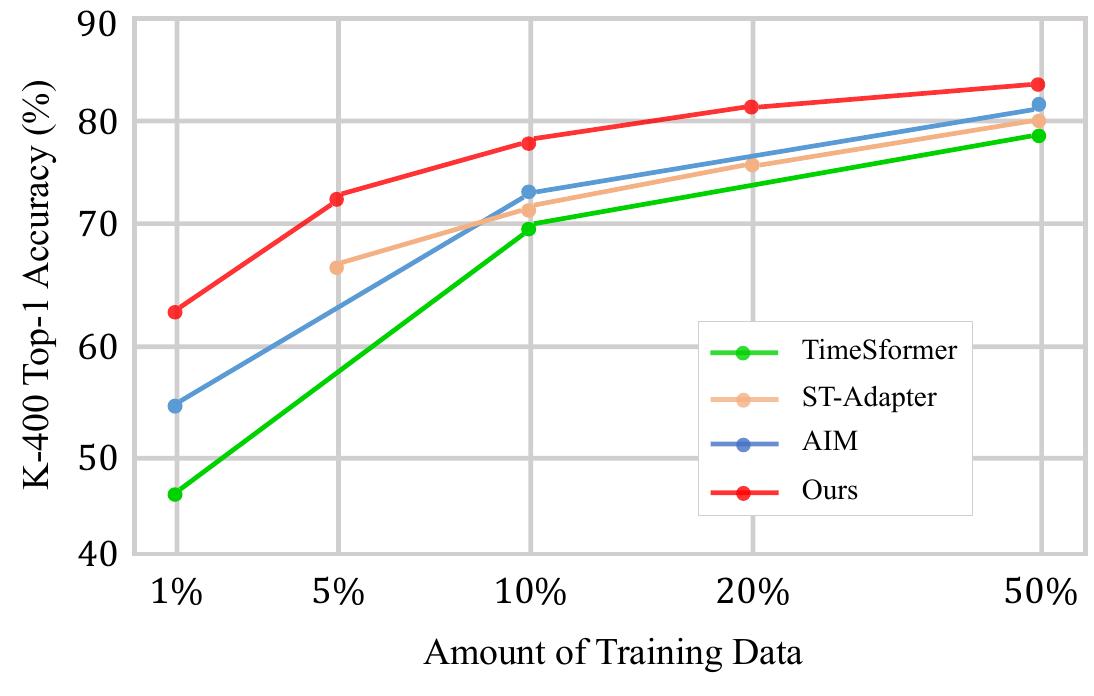}
        \caption{Comparison on training data efficiency. Our method demonstrates advantages especially with a small amount of training data.}
        \label{efficiency}
    \end{minipage}
\end{figure}

\noindent\textbf{Training Losses.}
We explore the effectiveness of our newly introduced object distillation loss $\mathcal{L}_{obj}$ and temporal reasoning loss $\mathcal{L}_{temp}$ in Table~\ref{loss}. We observe that $\mathcal{L}_{obj}$ leads to dramatic improvements on all three datasets, K-400, SSv2, and DAVIS-2017-UVOS. This is because $\mathcal{L}_{obj}$ offers rich semantic knowledge reference from the image foundation model to facilitate object identification. And better object decomposition further enhances high-level action recognition. As for $\mathcal{L}_{temp}$, it guides the model to capture the state changes on each object. Without this guidance, relying solely on the action classification loss is insufficient to direct the temporal module to perceive representative temporal dynamics, thus resulting in a drastic performance drop on the motion-heavy dataset SSv2. And notably, this temporal loss also leads to improvements on DAVIS, which verifies that the temporal perception in return bolsters object decomposition.

\begin{table}[]
\begin{minipage}{0.48\linewidth}
    \centering
    \small
    \caption{Ablation studies on training loss items $\mathcal{L}_{obj}$ and $\mathcal{L}_{temp}$. Note that $\mathcal{L}_{cls}$ is always applied in training.}
    \setlength{\tabcolsep}{1.5mm}{
    \begin{tabular}{ccccc}
    \toprule
        $\mathcal{L}_{obj}$ & $\mathcal{L}_{temp}$ & K-400 & SSv2 & DAVIS \\
        \midrule
        \xmark & \xmark & 70.4 & 44.8 & 27.3 \\
        \checkmark & \xmark & 80.3 & 55.2 & 39.5 \\
        \xmark & \checkmark & 73.5 & 52.1 & 30.5 \\
        \checkmark & \checkmark & 86.1 & 71.6 & 44.7 \\
    \bottomrule
    \end{tabular}}
    \label{loss}
\end{minipage}
\begin{minipage}{0.48\linewidth}
    \centering
    \small
    \caption{Ablation studies on the number of frames $T$ and sampling interval $\delta$.}
    \setlength{\tabcolsep}{1.5mm}{
    \begin{tabular}{ccccc}
    \toprule
        $T$ & $\delta$ & GFLOPs & K-400 & SSv2 \\
        \midrule
        4 & 1 & 4.2 & 85.4 & 70.7 \\
        8 & 2 & 8.4 & 85.8 & 71.2 \\
        8 & 4 & 8.2 & 85.8 & 71.1 \\
        16 & 2 & 18.0 & 86.1 & 71.6 \\
        16 & 4 & 17.8 & 86.1 & 71.6 \\
        \bottomrule
    \end{tabular}}
    \label{frame}    
\end{minipage}
\end{table}

% \noindent\textbf{Temporal Fusion Layer.}
% We study the structure of the early temporal fusion module ${\rm \Phi}_{\text{temp}}$ in Table~\ref{temp}. We vary the number of the temporal convolution layers in ${\rm \Phi}_{\text{temp}}$, where 0 means we instantiate ${\rm \Phi}_{\text{temp}}$ as identity mapping. The results indicate that one temporal convolution layer is sufficient for both video object segmentation and action recognition. This is because ${\rm \Phi}_{\text{temp}}$ is majorly used to exploit temporal dependency in consecutive frames to assist object decomposition, instead of action recognition. It does not require complex temporal modeling or large reception field, and simply aggregating information from adjacent frames is sufficient. 

\noindent\textbf{Temporal Sampling Strategy.}
We also study the effect of different temporal sampling strategies in training. Specifically, we vary the number of frames per clip $T$ and the time interval for object state change modeling $\delta$. In Table~\ref{frame}, we display the Top-1 accuracy on K-400 and SSv2, as well as the computation GFLOPs excluding ViT backbone for more intuitive comparison. An exciting phenomenon is that even on SSv2 that requires strong temporal modeling, we achieve promising results with very sparse sampling, i.e., only 4 frames as input with significantly reduced computation. Our method is generally robust to the temporal sampling hyper-parameters, demonstrating the effectiveness of our temporal reasoning on compressed object tokens.

\noindent\textbf{Training Data Efficiency.}
Fig.~\ref{efficiency} presents K-400 action recognition performance with various amounts of training data. For fair comparison with fully finetuned TimeSformer~\cite{bertasius2021space} as well as the efficient tuning methods, ST-Adapter~\cite{pan2022st} and AIM~\cite{yang2023aim}, we use CLIP ViT-B/16 with an input of 8 frames. Our method exhibits superior data efficiency to existing works. And the advantage becomes more significant with less training data, demonstrating the robust capacity of our object-centric formulation to transfer image pre-trained knowledge to videos.

\section{Conclusion}
In this work, we propose an object-centric formulation to adapt the image pre-trained models to the video domain. Specifically, we employ a frozen image pre-trained model to extract frame-wise features and use slot attention with learnable queries to parse each frame into a compact set of object tokens. Then we apply object-time interaction to these object tokens to explicitly establish object state changes across time span.
By using the distilled object tokens as the compressed representations of video frames, we have reduced computation costs and improved temporal reasoning capabilities especially under sparse sampling scenarios. Our method demonstrates state-of-the-art performance across action recognition benchmarks while realizing robust zero-shot video object segmentation, signifying a promising direction for future research in video analysis.

% In this work, we propose an object-centric formulation to adapt the image pre-trained models to video domain. Specifically, we employ a frozen image pre-trained model to extract frame-wise features, and use slot attention with learnable queries to parse each frame frame into a compact set of object tokens. Then we apply divided object-time attention on these object tokens to introduce object interactions and temporal reasoning, which significantly reduces computation redundancy. Our method achieves promising results on zero-shot video object segmentation and video action recognition with substantially higher efficiency and fewer tunable parameters. Further, our formulation is applicable to distinct image pre-training strategies and has the potential to leverage various large foundation models. However, there also exists a limitation that our object tokens majorly decompose different object semantics instead of instances. It might be more promising to utilize other modalities, e.g., optical flow, to assist more fine-grained object identification and further improve the knowledge transfer from images to videos.

\noindent\textbf{Limitations and Social Negative Impact.}
Our work demonstrates that the object-centric formulation serves as a reliable compression of video frames, which reduces redundancy and enables efficient temporal perception. But in some applications like a video QA system, it would be better to distill useful information according to specific questions. Incorporating questions as a condition to slot attention might address this problem. And there is a potential negative impact in privacy concerns. This work involves processing large amounts of visual data, which could include sensitive information and raise privacy concerns.

\section*{Acknowledgements}
This work is supported by GRF 14205719, TRS T41-603/20-R, Centre for Perceptual and Interactive Intelligence, and CUHK Interdisciplinary AI Research Institute.

\clearpage  % TODO REVIEW/FINAL: This \clearpage needs to be removed from both review and camera-ready versions.

\bibliographystyle{splncs04}
\bibliography{main}

\appendix

\section{More Experimental Results}
\noindent\textbf{Number of Object Tokens.}
We explore different number of object tokens $N$ in terms of the performance on action recognition as well as object discovery. For more intuitive comparison, we also exhibit the total number of tokens for temporal modeling, computation cost and the number of tunable parameters. The results are presented in Table~\ref{tab:token}. We use the frozen CLIP ViT-B/16 as $\rm \Phi_{\text{enc}}$, and report the results with $T=8$ frames per clip. Note that in the first row, using only $N=1$ object token means our method degenerates into spatially average pooling the frame-wise feature maps into vectors of size $\mathbb{R}^{T\times D}$ to do temporal reasoning. And in the last row, `None' means we keep the original features of size $\mathbb{R}^{T\times HW\times D}$ for temporal modeling. These two settings do not require the slot attention module for object decomposition, thus resulting in fewer tunable parameters. The total number of the tokens used for temporal reasoning equals to $T\times N$. From the comparison, we have three observations. (1) Our method significantly outperforms the average pooling baseline with negligible computation overhead. (2) More object tokens lead to more fine-grained object decomposition, i.e., higher results on DAVIS. And the action recognition performance maintains stable when $N\geq 8$. Hence, we take $N=8$ object tokens in default as a good choice for the trade-off between performance and efficiency. (3) Despite that our method requires much fewer computation GFLOPs than keeping the original feature for temporal modeling, we still achieve significantly superior performance on action recognition benchmarks. This is because the original video features contain much redundant information, which results in meaningless noise for state change modeling. In contrast, we compress the videos into meaningful object tokens, providing more intuitive and interpretable cues to identify temporal dynamics.

\noindent\textbf{Temporal Modeling Strategy.}
We also explore how different time intervals $\delta$ in the temporal state change modeling influences the performance in Table~\ref{tab:layer}. The comparisons are based on CLIP ViT-B/16 with an input of $T=8$ frames. We present different $\delta$ value with its corresponding number of state change vectors per object, denoted as $T'=T-\delta$. Note that in the last row, `All' means we traverse all different $\delta=\{1,2,...,T-1\}$ and combine the state change vectors calculated from all time intervals, thus resulting in $T'=\frac{T(T-1)}{2}$. We observe that when the time interval $\delta$ increases, the action recognition performance first improves, then maintains stable and eventually drops. This is because too small interval may result in very subtle state changes in adjacent time stamps, thus resulting in non-representative state change modeling. And too large interval could make the model skip some frames, e.g., when $\delta>\frac{T}{2}$, the frames with time stamp $\frac{T}{2}<t<1+\delta$ will be ignored in the temporal modeling stage. This leads to lossy temporal reasoning and causes the performance drop. Another observation is that it is unnecessary to traverse all different kinds of combinations, simply taking $\delta=2$ is sufficient in this condition.

\begin{table*}[t]
    \centering
    \setlength{\tabcolsep}{2mm}{
    \begin{tabular}{ccccccc}
    \toprule
        $N$ & Total Tokens & GFLOPs & Tunable Param & K-400 & SSv2 & DAVIS \\
        \midrule
        1 & 8 & 273.4 & 1.6 & 72.5 & 44.1 & - \\
        4 & 32 & 279.8 & 6.5 & 83.1 & 67.5 & 37.5 \\
        8 & 64 & 281.2 & 6.5 & 85.8 & 71.2 & 42.7 \\
        16 & 128 & 282.7 & 6.5 & 85.9 & 71.2 & 43.0 \\
        None & 2048 & 314.2 & 4.1 & 81.1 & 64.3 & - \\
        \bottomrule
    \end{tabular}}
    \caption{Ablation on the number of object tokens. We report Top-1 action recognition accuracy on K-400 and SSv2, and $\mathcal{J}\&\mathcal{F}$ score on multiple object segmentation DAVIS-2017-Unsupervised.}
    \label{tab:token}
\end{table*}

\begin{table*}[t]
    \centering
    \setlength{\tabcolsep}{2mm}{
    \begin{tabular}{cccccc}
    \toprule
        $\delta$ & $T'$ & GFLOPs & K-400 & SSv2 & DAVIS \\
        \midrule
        1 & 7 & 281.3 & 85.5 & 70.4 & 42.7 \\
        2 & 6 & 281.2 & 85.9 & 71.2 & 42.7 \\
        3 & 5 & 281.1 & 85.9 & 71.1 & 42.6 \\
        4 & 4 & 281.0 & 85.7 & 70.9 & 42.7 \\
        5 & 3 & 280.9 & 85.5 & 70.5 & 42.5 \\
        6 & 2 & 280.8 & 85.2 & 69.6 & 42.4 \\
        7 & 1 & 280.7 & 85.1 & 69.3 & 42.4 \\
        All & 28 & 283.4 & 85.9 & 71.2 & 42.6 \\
        \bottomrule
    \end{tabular}}
    \caption{Ablation on the temporal modeling formulation. We report Top-1 action recognition accuracy on K-400 and SSv2, and $\mathcal{J}\&\mathcal{F}$ score on multiple object segmentation DAVIS-2017-Unsupervised.}
    \label{tab:layer}
\end{table*}

\begin{table*}[]
    \centering
    \setlength{\tabcolsep}{2mm}{
    \begin{tabular}{lccccc}
    \toprule
        Method & GPUs & Pre-train & Training Hours & Total GPU Hours & Ratio \\
        \midrule
        TimeSformer~\cite{bertasius2021space} & 8 & IN-21K & 71 & 568 & 3.0$\times$ \\
        ST-Adapter~\cite{pan2022st} & 8 & CLIP & 37 & 296 & 1.5$\times$ \\
        AIM~\cite{yang2023aim} & 8 & CLIP & 39 & 312 & 1.6$\times$ \\
        EVL~\cite{lin2022frozen} & 8 & CLIP & 35 & 280 & 1.5$\times$ \\
        Ours & 8 & CLIP & 24 & 192 & 1.0$\times$ \\
        \bottomrule
    \end{tabular}}
    \caption{Comparison on training consumption. We compare with both fully fine-tuned and efficient tuning methods and report the training GPU hours.}
    \label{tab:training}
\end{table*}

\begin{table}[]
    \centering
    \setlength{\tabcolsep}{2mm}{
    \begin{tabular}{lcc}
    \toprule
        Method & Throughput & Ratio \\
        \midrule
        ST-Adapter~\cite{pan2022st} & 25 & 0.8$\times$ \\
        AIM~\cite{yang2023aim} & 23 & 0.8$\times$ \\
        EVL~\cite{lin2022frozen} & 18 & 0.6$\times$ \\
        Ours & 30 & 1.0$\times$ \\
        \bottomrule
    \end{tabular}}
    \caption{Comparison on inference speed. We compare with typical efficient tuning counterparts and report the inference throughput in clips per second.}
    \label{tab:inference}
\end{table}

\noindent\textbf{Training and Inference Efficiency.}
Additionally, we also explore the training and inference efficiency of our method. For training, we compare with fully fine-tuned TimeSformer~\cite{bertasius2021space} as well as efficient tuned ST-Adapter~\cite{pan2022st}, AIM~\cite{yang2023aim} and EVL~\cite{lin2022frozen}. For fair comparison, the backbone of each method is built on ViT-B/16. We run the training process on the same 8 NVIDIA 3090 GPU server, and report the training hours and total GPU hours. Based on our reimplementations, the results are shown in Table~\ref{tab:training}. First, the fully fine-tuned TimeSformer takes far more training hours to converge due to much more tunable parameters. Second, the efficient tuning methods, ST-Adapter, AIM and EVL, respectively takes around 1.4x, 1.5x, 1.3x training time than ours. This is because ST-Adapter and AIM have tunable parameters in shallow layers that require additional time in backpropagation, and EVL is equipped with a heavy decoder to deal with temporal dynamics. In contrast, our object-centric formulation achieves image-to-video adaptation without requiring redundant temporal modeling or modifying the backbone parameters, which leads to higher training efficiency. For inference, we present the throughput comparison in Table~\ref{tab:inference}. Our method reaches the highest throughput among the efficient tuning counterparts, demonstrating the efficiency of our object-centric temporal reasoning strategy.

\begin{table}[]
    \centering
    \setlength{\tabcolsep}{2mm}{
    \begin{tabular}{lcc}
    \toprule
        Training & Foreground & Background \\
        \midrule
        K-400 & 20.58 & 1.79 \\
        SSv2 & 25.63 & 3.61 \\
        \bottomrule
    \end{tabular}}
    \caption{Comparison on the $\mathcal{L}_2$ norm of state change vectors. We compare the norm values of tokens corresponding to foreground objects and background areas respectively based on K-400 and SSv2 trained models.}
    \label{norm}
\end{table}

\noindent\textbf{Statistics on State Change Vectors.}
Considering that each video only contains a small portion of object semantics, some tokens only attend to background areas as verified in Fig.~\ref{att}. To this end, we further investigate whether the object tokens that focus on backgrounds influence the temporal modeling process. To achieve this goal, we utilize the object annotations in DAVIS-2017-Unsupervised to discriminate foreground and background tokens, and make statistics on the $\mathcal{L}_2$ norm of the state change vectors. Specifically, we use the model with ViT-B/16 backbone respectively trained on K-400 and SSv2 for analysis. Following the standard evaluation pipeline of DAVIS-2017-Unsupervised, in each video, we denote the object tokens that are assigned to ground truth objects as foreground tokens, and the rest as background tokens. Next, we calculate the $\mathcal{L}_2$ norm of the state change vectors of all object tokens, and compute the respective averages for foreground and background tokens. We present the statistic results in Table~\ref{norm}. It is evident that the state change vectors of foreground objects dominate over the tokens corresponding to backgrounds. Therefore, the average pooling operation over all state change vectors is approximately equivalent to averaging the state changes of the foreground objects, suppressing the interference of background areas.

\begin{table}[]
    \centering
    \caption{Ablation studies on object-time interaction design. We compare different combinations of our object interaction module $\rm \Phi_{oi}$, temporal interaction module $\rm \Phi_{ti}$ and an alternative temporal attention module $\rm \Phi_{ta}$.}
    \setlength{\tabcolsep}{2mm}{
    \begin{tabular}{cccccccc}
    \toprule
        $\rm \Phi_{oi}$ & $\rm \Phi_{ti}$ & $\rm \Phi_{ta}$ & GFLOPs & Tunable Param & K-400 & SSv2 & DAVIS \\
        \midrule
        \xmark & \xmark & \xmark & 13.7 & 2.4 & 77.5 & 49.1 & 36.4 \\
        \checkmark & \xmark & \xmark & 15.9 & 4.8 & 79.1 & 50.0 & 39.4 \\
        \xmark & \checkmark & \xmark & 15.6 & 4.0 & 83.1 & 63.9 & 41.7 \\
        \xmark & \xmark & \checkmark & 23.9 & 5.1 & 80.2 & 57.8 & 40.0 \\
        \checkmark & \checkmark & \xmark & 17.8 & 6.5 & 86.1 & 71.6 & 44.7 \\
        \checkmark & \xmark & \checkmark & 26.1 & 7.5 & 85.4 & 66.8 & 43.5 \\
        \bottomrule
    \end{tabular}}
    \label{objtime}
\end{table}

\noindent\textbf{Object-Time Interaction Architecture.}
Additionally, we also present the ablation study on the detailed object-time interaction module design. Besides the interaction modules, $\rm \Phi_{oi}$ and $\rm \Phi_{ti}$, introduced in this paper, we also compare with prevalent temporal attention module $\rm \Phi_{ta}$ instantiated in the form of TimeSformer~\cite{lin2020space}. We compare different settings in terms of the computation GFLOPs excluding ViT backbone, the number of tunable parameters and downstream task performance. From Table~\ref{objtime}, we conclude three major observations. First, the object interaction module is necessary since some actions or events require an understanding of the relationships between contextual objects. Second, the temporal interaction not only enhances action recognition but also in return facilitates object identification. Third, our temporal modeling in the form of object state changes surpasses the temporal attention operation both in performance and efficiency. We conjecture this is because we have parsed video frames into separated object tokens. Compared to densely fusing the object representations in the form of attention, explicitly modeling object state changes between sparsely sampled frames provides more intuitive cues for video understanding.

\section{More Visualization Results}
\begin{figure}
    \centering
    \includegraphics[width=0.23\linewidth]{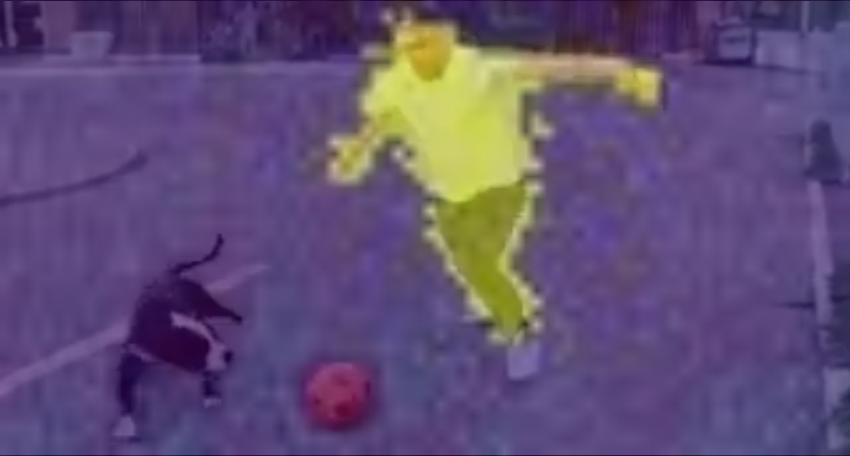}
    \includegraphics[width=0.23\linewidth]{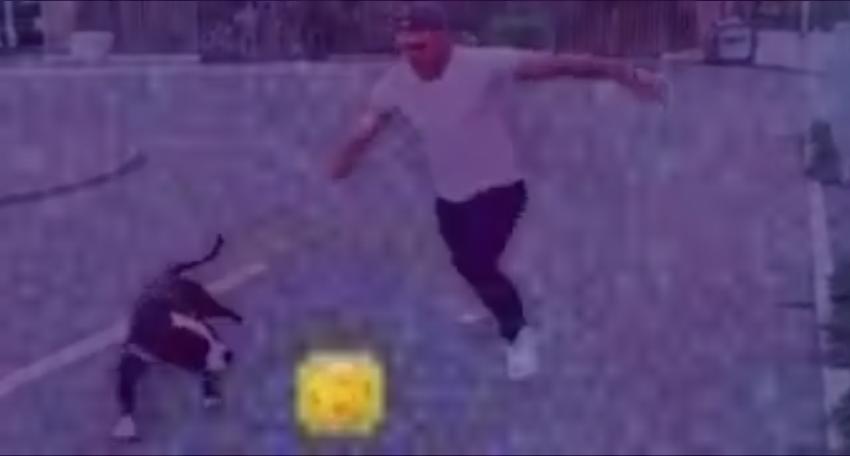}
    \includegraphics[width=0.23\linewidth]{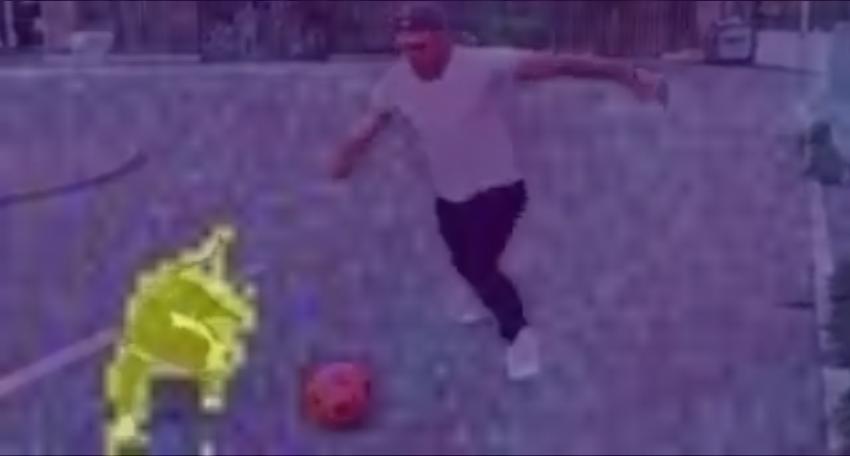}
    \includegraphics[width=0.23\linewidth]{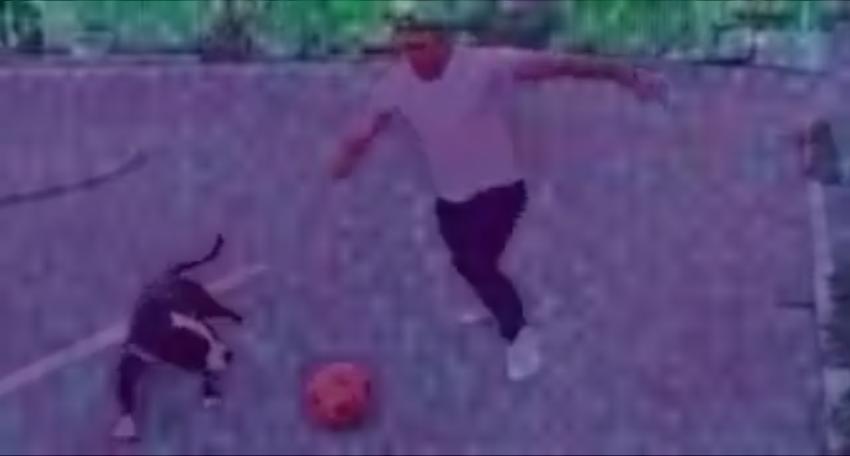}
    \includegraphics[width=0.23\linewidth]{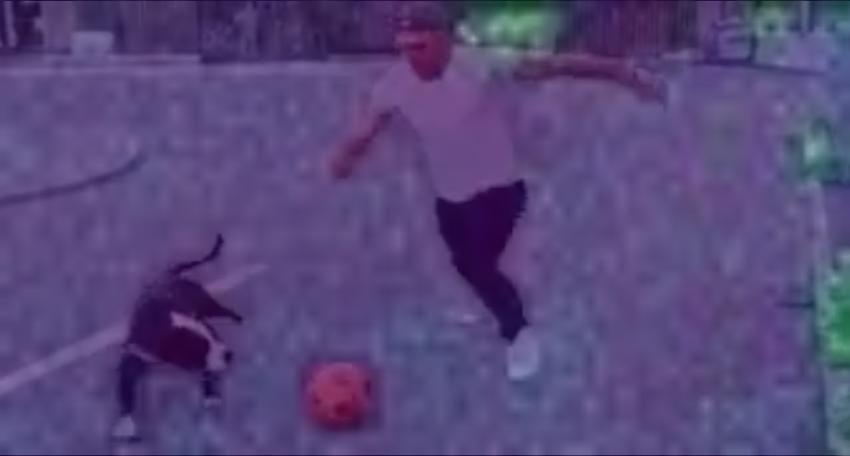}
    \includegraphics[width=0.23\linewidth]{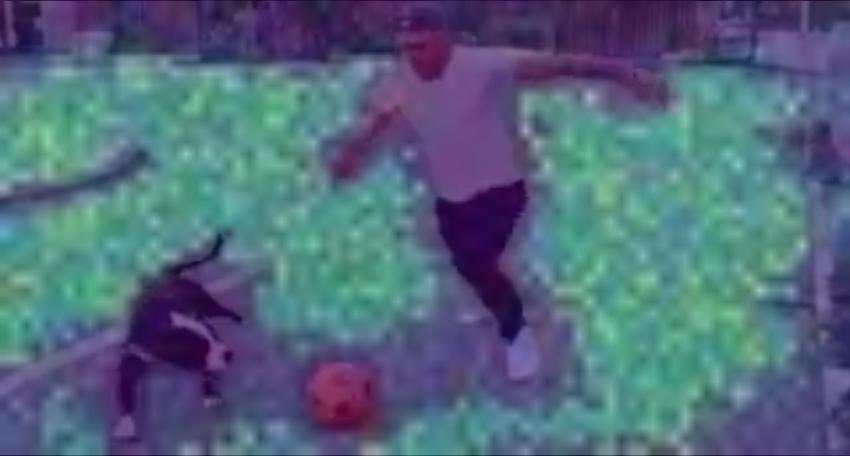}
    \includegraphics[width=0.23\linewidth]{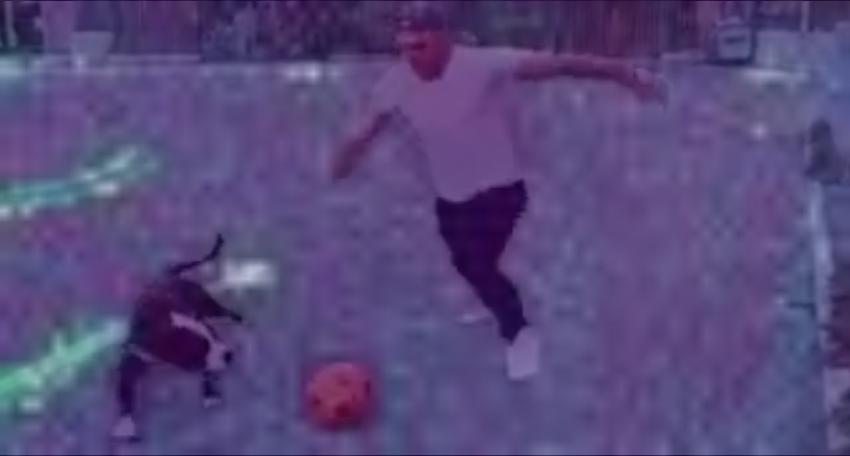}
    \includegraphics[width=0.23\linewidth]{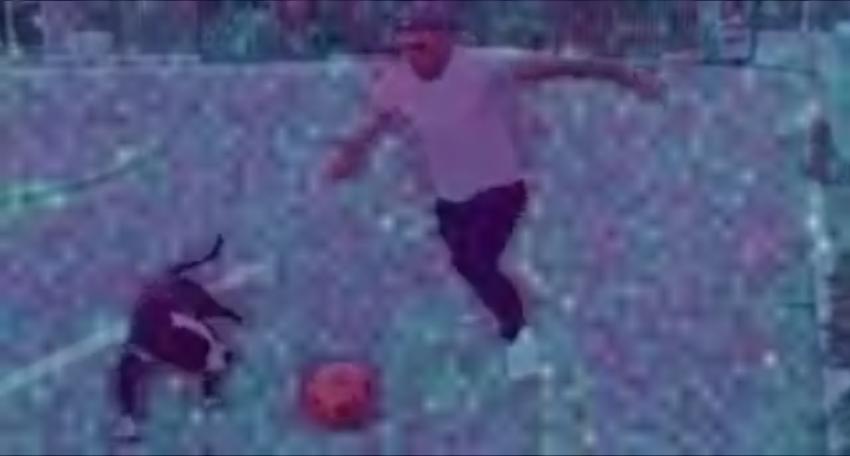}
    \caption{Visualization of the slot attention maps of all $N=8$ object tokens. We present the results without binarization.}
    \label{att}
\end{figure}

\noindent\textbf{Visualization of All Tokens.}
We first show the soft attention masks of all $N=8$ object tokens in Fig.~\ref{att}. There are three tokens respectively presenting salient activations on three primary objects, human, dog, and soccer ball. The rest five tokens attend to different background areas with low activations. It aligns with the property of slot attention, where the foreground objects are separated into different groups and the background elements tend to be evenly distributed among the rest slots.

\begin{figure*}
    \centering
    \begin{minipage}{\textwidth}
        \centering
        \includegraphics[width=\linewidth]{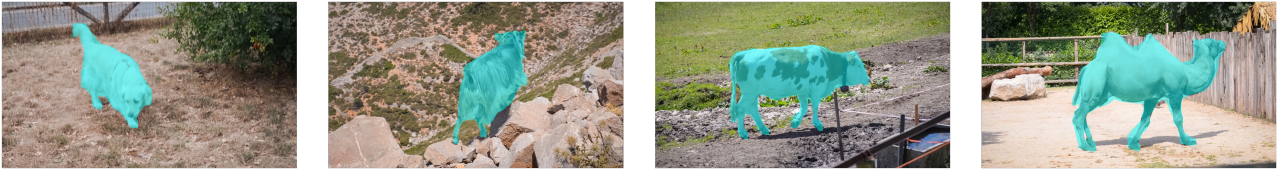}
        \caption*{Object discovery results trained on K-400.}
        \label{fig:k400}
    \end{minipage}%

    \begin{minipage}{\textwidth}
        \centering
        \includegraphics[width=\linewidth]{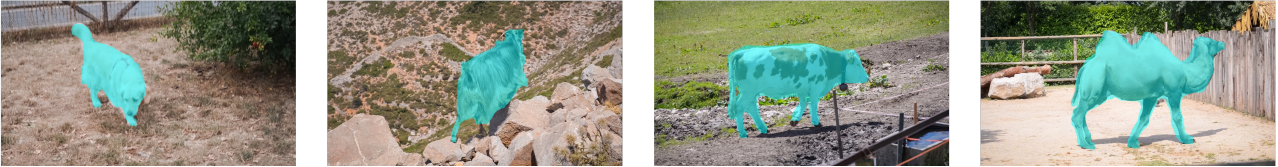}
        \caption*{Object discovery results trained on SSv2.}
        \label{fig:ssv2}
    \end{minipage}
    \caption{Visualization of the segmentation results with different training data. We present the results on a set of scenes with quadruped animals. We compare the object segmentation results with the model trained on K-400 and SSv2 respectively.}
    \label{fig:unseen}
\end{figure*}
\noindent\textbf{Generalization to Unseen Scenarios.}
In our formulation, we aim to distill general object knowledge from image pre-trained models into the object tokens to discover different object components in videos. Here we verify whether the distilled object tokens can generalize to unseen scenarios with novel objects. To do this, we compare our models trained on SSv2 and K-400 with only self-supervised objectives, $\mathcal{L}_{obj}$ and $\mathcal{L}_{temp}$, in Fig.~\ref{fig:unseen}. We present the object discovery results on a series of scenes with quadruped animals, e.g., dog, goat, camel. Note that K-400 contains these objects while SSv2 does not contain these object semantics. From the comparison, we observe that the learned object tokens are able to discover these animals regardless of the training data with the only difference in some borders. The promising segmentation results with SSv2 trained model demonstrates the generalization ability to unseen objects in a zero-shot manner. This phenomenon also verifies that the object tokens are able to transfer general object knowledge from image pre-trained models, rather than overfitting the training samples. 

\end{document}